\documentclass[letterpaper]{article} 
\usepackage{aaai2026}  
\usepackage{times}  
\usepackage{helvet}  
\usepackage{courier}  
\usepackage[hyphens]{url}  
\usepackage{graphicx} 
\usepackage{natbib}  
\usepackage{caption} 
\usepackage{algorithm}
\usepackage{algorithmic}

\usepackage{amsmath}
\usepackage{amssymb}
\usepackage{algorithm}
\usepackage{algorithmic}
\usepackage{multirow}
\usepackage{diagbox}
\usepackage{adjustbox}
\usepackage{tabularx}
\usepackage{rotating}
\usepackage{booktabs}
\usepackage{xcolor}
\usepackage{colortbl}
\usepackage{xr}
\usepackage{url}

\usepackage{color}
\usepackage{xcolor,colortbl}
\usepackage{mathtools}
\usepackage{bm} 
\usepackage{pifont}
\usepackage{subcaption} 
\usepackage{enumitem} 
\usepackage{amsfonts}
\usepackage{booktabs}
\usepackage{xspace}
\usepackage{url}
\usepackage{newfloat}
\usepackage{listings}
\DeclareCaptionStyle{ruled}{labelfont=normalfont,labelsep=colon,strut=off} 
\floatstyle{ruled}
\newfloat{listing}{tb}{lst}{}
\floatname{listing}{Listing}
\newcommand{\hide}[1]{}

\definecolor{purple}{rgb}{0.65,0,0.65}
\definecolor{dark_green}{rgb}{0, 0.5, 0}
\definecolor{blueish}{rgb}{0.0, 0.3, .6}
\definecolor{LightCyan}{rgb}{0.88,0.95,1}
\definecolor{LightGrey}{rgb}{0.56,0.56,0.56}
\definecolor{tabhighlight}{HTML}{f2f0f0} 
\newcolumntype{h}{>{\columncolor{tabhighlight}}c}
\definecolor{positivetabledelta}{rgb}{0.172, 0.750, 0.098} 
\definecolor{negativetabledelta}{rgb}{0.851, 0.2, 0.078}

\newcommand{\ie}{\textit{i.e.}}       
\newcommand{\eg}{\textit{e.g.}}       
\newcommand{\etal}{\textit{et al.}}    
       
\newcommand{\vs}{\textit{vs.}}

\newcommand{\srcc}{SRCC\xspace}
\newcommand{\plcc}{PLCC\xspace}

\title{Beyond Cosine Similarity: Magnitude-Aware CLIP for No-Reference Image Quality Assessment}
\author{
    Zhicheng Liao\textsuperscript{\rm 1},
    Dongxu Wu\textsuperscript{\rm 1},
    Zhenshan Shi\textsuperscript{\rm 1},
    Sijie Mai\textsuperscript{\rm 1},
    Hanwei Zhu\textsuperscript{\rm 2},
    Lingyu Zhu\textsuperscript{\rm 3}, \\
    Yuncheng Jiang\textsuperscript{\rm 1},
    Baoliang Chen\textsuperscript{\rm 1}\thanks{Corresponding author.}
}
\affiliations{
    \textsuperscript{\rm 1}School of Computer Science, South China Normal University, China\\
    \textsuperscript{\rm 2}School of Computer Science and Engineering, Nanyang Technological University, Singapore\\
    \textsuperscript{\rm 3}School of Computer Science, City University of Hong Kong, China\\
    zcliao@m.scnu.edu.cn, blchen@m.scnu.edu.cn
}

\usepackage{bibentry}

\begin{document}

\maketitle

\begin{abstract}
Recent efforts have repurposed the Contrastive Language-Image Pre-training (CLIP) model for No-Reference Image Quality Assessment (NR-IQA) by measuring the cosine similarity between the image embedding and textual prompts such as ``a good photo” or ``a bad photo.” However, this semantic similarity overlooks a critical yet underexplored cue: \textbf{\textit{the magnitude of the CLIP image features, which we empirically find to exhibit a strong correlation with perceptual quality.}}
In this work, we introduce a novel adaptive fusion framework that complements cosine similarity with a magnitude-aware quality cue. Specifically, we first extract the absolute CLIP image features and apply a Box-Cox transformation to statistically normalize the feature distribution and mitigate semantic sensitivity. The resulting scalar summary serves as a semantically-normalized auxiliary cue that complements cosine-based prompt matching. To integrate both cues effectively, we further design a confidence-guided fusion scheme that adaptively weighs each term according to its relative strength.
Extensive experiments on multiple benchmark IQA datasets demonstrate that our method consistently outperforms standard CLIP-based IQA and state-of-the-art baselines, \textbf{\textit{without any task-specific training}}. 

\end{abstract}

\begin{links}
    \link{Code}{https://github.com/zhix000/MA-CLIP}
\end{links}

\section{Introduction}
\begin{figure}[t]
  \centering
      \includegraphics[width=\linewidth]{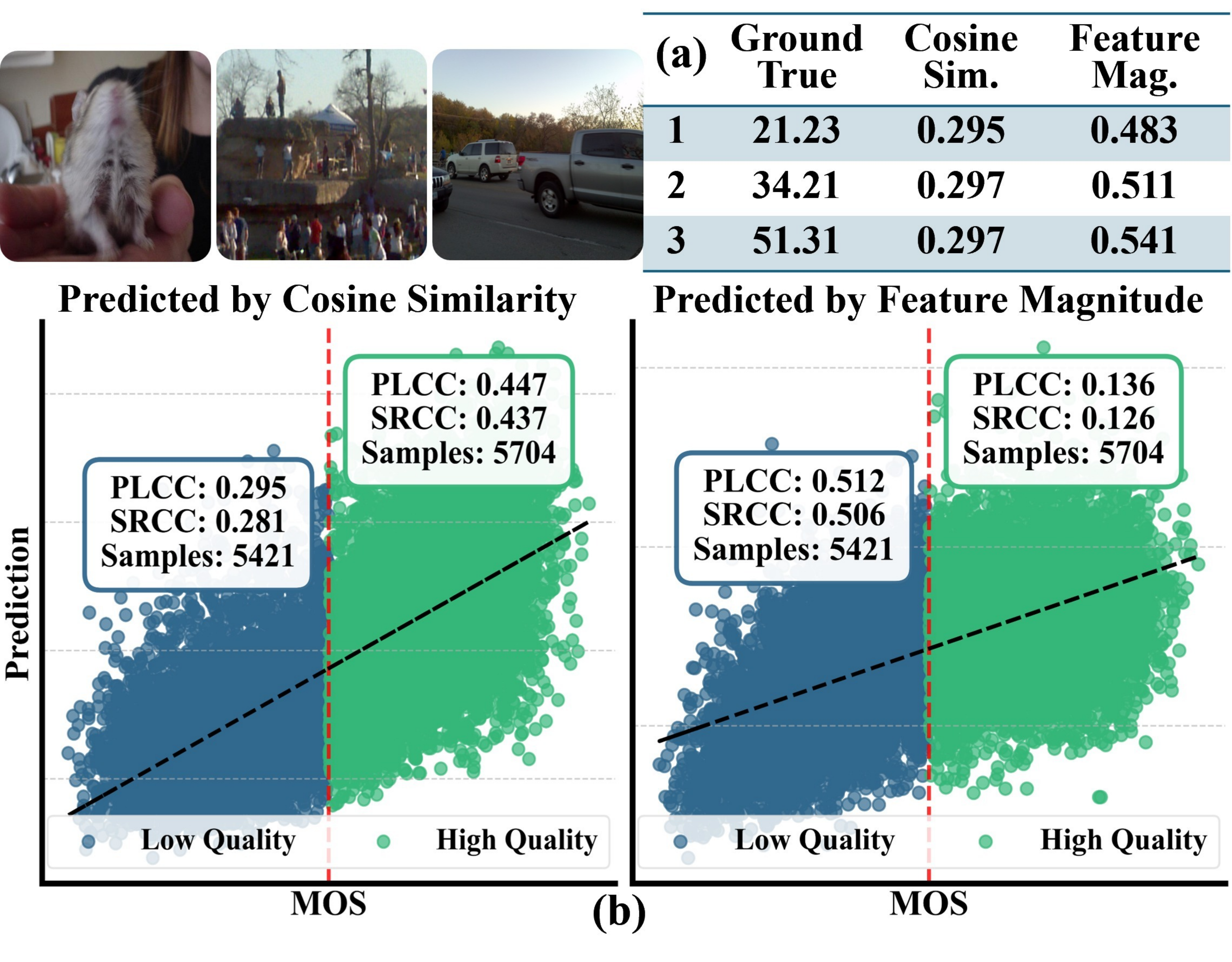}
      \vspace{-22pt}
  \caption{
  \textbf{(a)} Limitations of prompt-based CLIP-IQA: although the images exhibit a wide range of perceptual quality (reflected in their MOS), the cosine similarity between the image embedding and textual prompts remains nearly constant. In contrast, the feature magnitude shows a strong correlation with MOS.  
  \textbf{(b)} Complementary behaviors of the two cues across quality levels: As the scatter plots of SPAQ dataset shows that cosine similarity is more reliable in the high-quality region, where semantic features align well with CLIP's pretrained distribution; feature magnitude is more discriminative under low-quality distortions, where semantic alignment breaks down.  These observations motivate our dual-cue fusion framework that adaptively integrates both signals for robust quality prediction.
  }
  \label{fig:motivation}
\end{figure}

Image Quality Assessment (IQA) aims to automatically predict the perceptual quality of an image, playing a critical role in a wide range of applications, such as image enhancement, compression, generation, and transmission. In practical scenarios where pristine reference images are unavailable, No-Reference IQA (NR-IQA) becomes particularly essential, as it assesses image quality solely based on the distorted image.

Recently, significant progress has been made in learning-based NR-IQA models. However, most existing approaches rely heavily on supervised training with specific IQA datasets, often overfitting to dataset-specific distortions or content. This generalization bottleneck severely limits their applicability in real-world systems. The emergence of large-scale vision-language models such as CLIP~\cite{radford2021learning} offers a promising alternative. Trained on hundreds of millions of image–text pairs, CLIP demonstrates remarkable generalization and semantic understanding in a zero-shot manner. Recent works have adapted CLIP for NR-IQA by leveraging its ability to compute the cosine similarity between image embedding and quality-descriptive textual prompts (\eg, ``a good photo'' \vs  ``a bad photo'')~\cite{wang2023exploring}. This prompt-based CLIP-IQA technique has shown surprising effectiveness on standard IQA benchmarks without any fine-tuning, providing a compelling direction for generic, training-free quality assessment.

However, CLIP-IQA relies solely on semantic prompt similarity, overlooking another critical yet underexplored signal inherent in the model. Specifically, the cosine similarity computation involves $\ell_2$ normalization of the image features, which removes the magnitude information entirely. Through our extensive empirical observations, we find that the magnitude (\ie, norm) of the CLIP image embedding, although ignored in standard usage, is in fact highly indicative of perceptual quality. As illustrated in Fig.~\ref{fig:motivation}(a), images with widely varying MOS often yield nearly identical prompt-based similarities, failing to capture true perceptual differences. In contrast, the feature magnitude varies consistently with MOS, increasing for higher-quality images and decreasing for lower-quality ones. Moreover, we observe that cosine-based scores are more reliable in distinguishing high-quality images, where semantic features remain well aligned with CLIP’s pretrained distribution, while magnitude cues are more sensitive and consistent in low-quality regimes, where distortions cause semantic misalignment (see Fig.~\ref{fig:motivation}(b)). This insight suggests a key conclusion: \textit{\textbf{cosine similarity and feature magnitude are complementary}}. Motivated by this, we propose to leverage both cues jointly rather than relying on either alone.

To this end, we introduce an adaptive dual-cue fusion framework that integrates semantic and magnitude information for more robust quality prediction. Specifically, we compute two scalar quality indicators: (1) the conventional cosine similarity between image embedding and textual quality prompts, and (2) a magnitude-based cue derived directly from the image features. For the latter, we first take the absolute value of each feature dimension and apply Box-Cox transformation to statistically normalize their distributions across images. This normalization mitigates semantic-content bias and aligns the magnitudes to near-Gaussian distribution. We then average the transformed values to obtain a stable, debiased magnitude score. Finally, to fully exploit their complementary strengths, we design a confidence-guided fusion mechanism that adaptively weights the two cues based on their estimated reliability. This allows the model to trust the cosine score more in high-quality conditions where image semantics are well recognizable, and rely more on the normalized magnitude cue under severe distortions where semantic similarity becomes less reliable.

Extensive experiments on multiple IQA benchmarks validate the effectiveness of our approach. Without any task-specific training, our method substantially outperforms the vanilla CLIP-IQA baseline and recent state-of-the-art NR-IQA models. 
These results highlight the benefits of combining semantic and magnitude cues for robust and accurate image quality prediction. Our main contributions are summarized as follows:

\begin{itemize}
  \item We identify the magnitude of CLIP image embedding as a strong and previously overlooked quality cue for NR-IQA that complements traditional cosine similarity.
  \item We introduce a Box-Cox transformation to normalize per-dimension embedding magnitudes, producing a statistically consistent quality indicator across diverse image contents.
  \item We design a confidence-guided fusion strategy that adaptively weights cosine similarity and magnitude cues based on their relative reliability.
  \item Our method is entirely training-free, achieves state-of-the-art zero-shot IQA performance, and generalizes effectively to different image content, demonstrating the versatility of the proposed dual-cue framework.
\end{itemize}

\begin{figure*}[t]
    \centering
    \includegraphics[width=1.0\linewidth]{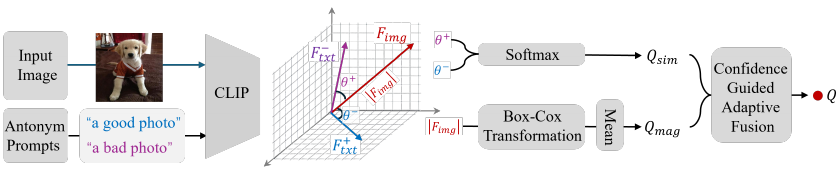}
    \caption{Overview of the Proposed Magnitude-Aware CLIP IQA Framework. Given an input image, we extract its CLIP image embedding and compute two quality signals: (1) $Q_{\text{sim}}$, the image semantic similarity with text prompts, and (2) $Q_{\text{mag}}$, a magnitude-based score obtained via Box-Cox transformation for statistical normalization. To balance these complementary cues, we adopt a confidence discrepancy and generate softmax-based fusion weights, producing the final quality prediction $Q$.}
    \label{fig:Overview}
\end{figure*}

\section{Related Work}
\subsection*{Vision Based NR-IQA Models}
Early no-reference IQA approaches relied on handcrafted features that captured natural scene statistics (NSS), with descriptors derived from spatial~\cite{mittal2012making}, wavelet~\cite{moorthy2010two}, and DCT~\cite{saad2012blind} domains. Psychovisual models inspired by the free-energy principle of the human visual system further enriched the modeling of perceptual degradation~\cite{gu2014using, zhai2011psychovisual}. With the advent of deep learning, CNN-based models became prevalent. Early works such as Kang~\etal~\cite{kang2014convolutional} learned quality-aware representations directly from image patches, and subsequent extensions introduced multi-task training~\cite{kang2015simultaneous} to jointly predict quality and distortion types. Later models such as DIQaM~\cite{bosse2017deep},  FPR~\cite{chen2020no}, and GraphIQA~\cite{sun2022graphiqa} further advanced performance by modeling spatial dependencies and incorporating relational reasoning. Transformer-based models have also been explored due to their strong modeling capacity and flexibility. You~\etal~\cite{you2021transformer} introduced a Transformer backbone for NR-IQA that benefits from features pretrained on large-scale classification tasks. Other approaches leverage perceptual priors from image restoration networks~\cite{lin2018hallucinated, chen2022no}.  However, these models struggle to generalize due to limited data and domain-specific bias. To address this, recent work incorporates auxiliary tasks~\cite{you2021transformer, lin2018hallucinated} and learning strategies such as meta-learning~\cite{zhu2020metaiqa}, curriculum learning~\cite{wang2023hierarchical}, and domain generalization~\cite{chen2021learning, chen2025monotonic}. Despite progress, generalization in NR-IQA remains an open problem.

\subsection*{Vision-Language Based NR-IQA Models}
Vision-language models provide a new paradigm for NR-IQA. Wang \etal~\cite{wang2023exploring} first leveraged CLIP~\cite{radford2021learning} to compute the similarity between distorted images and antonym-paired textual prompts. Follow-up work such as IPCE~\cite{peng2024aigc} mapped cosine similarity scores to discrete quality levels using hand-crafted prompts, while CLIP-AGIQA~\cite{tang2024clip} learned prompt tokens for six quality classes and concatenated them with image features for quality regression. In parallel, benchmarks such as Q-Bench~\cite{wu2023qbench} and DepictQA~\cite{you2024descriptive} evaluated the IQA capability of large multimodal models (LMMs) on low-level visual perception. Zhu \etal~\cite{zhu20242afc} examined 2AFC-style prompting for preference judgment, and Q-Align~\cite{wu2023align} proposed rating-level prompts to elicit more consistent predictions. Further improvements have been achieved by leveraging pre-trained LMMs and high-quality instruction datasets~\cite{wu2024towards, zhu2024adaptive}. However, fine-tuning these models on IQA tasks often results in catastrophic forgetting~\cite{luo2023empirical}, weakening performance on other domains. This motivates the need to explore training-free strategies that better preserve the generality of pre-trained vision-language models. 

\section{Method}
\subsection{Preliminary: Semantic Similarity Based CLIP-IQA}
\label{subsec:prelim}
In the classical CLIP-IQA model, the zero-shot capability of the CLIP model for NR-IQA  has been explored. In particular, the image quality is estimated by measuring the alignment between image embedding and handcrafted quality prompts, such as \{\textit{``a good photo''}\} \vs \{\textit{``a bad photo''}\}. Formally, let $\mathbf{x} \in \mathbb{R}^{H \times W \times 3}$ denote an input image, and let $\mathcal{T}_{{pos}}$ and $\mathcal{T}_{{neg}}$ denote a pair of antonymic textual descriptions reflecting high and low perceptual quality, respectively. The CLIP model is utilized to encode the image and text into embedding vectors:
\begin{align}
    {F}_{img} &= \phi_{{img}}(\mathbf{x}) \in \mathbb{R}^{D}, \\
    {F}_{txt}^{+} &= \phi_{{txt}}(\mathcal{T}_{{pos}}) \in \mathbb{R}^{D}, \\
    {F}_{txt}^{-} &= \phi_{{txt}}(\mathcal{T}_{{neg}}) \in \mathbb{R}^{D},
\end{align}
where $\phi_{{img}}(\cdot)$ and $\phi_{{text}}(\cdot)$ are the image and text encoders of the pre-trained CLIP model, respectively. The cosine similarity between the normalized embeddings is computed as:
\begin{align}
\label{eqn:cos}
    s^+ &= \cos({\mathbf{}}F_{img},{{F}}_{text}^{+}) = {\hat{{F}}_{img} \cdot \hat{{F}}_{text}^{+}},\\
    s^- &= \cos({{F}}_{img},{{F}}_{text}^{-}) = {\hat{{F}}_{img} \cdot \hat{{F}}_{text}^{-}},
\end{align}
where $\hat{{F}} = {F} / \|{F}\|_2$, denotes the $\ell_2$-normalized results. The final quality score is obtained using a softmax-based probability:
\begin{equation}
    Q_{{sim}} = \frac{\exp(s^+ / \tau)}{\exp(s^+ / \tau) + \exp(s^- / \tau)},
\end{equation}
where $\tau$ is a temperature hyperparameter. The  $Q_{{sim}}$ reflects the relative probability that the image with good quality than the bad one, which can be deemed as the image quality assessment result.

\subsection{Magnitude-Aware CLIP IQA Model}
An overview of the proposed framework is illustrated in Fig.~\ref{fig:Overview}. The pipeline are described in detail as follows.
\label{subsec:magnitude-aware}

\subsubsection{Limitations of Cosine Similarity}
\label{subsubsec:limitation}

\begin{figure}[t]
    \centering
    \includegraphics[width=1.0\linewidth]{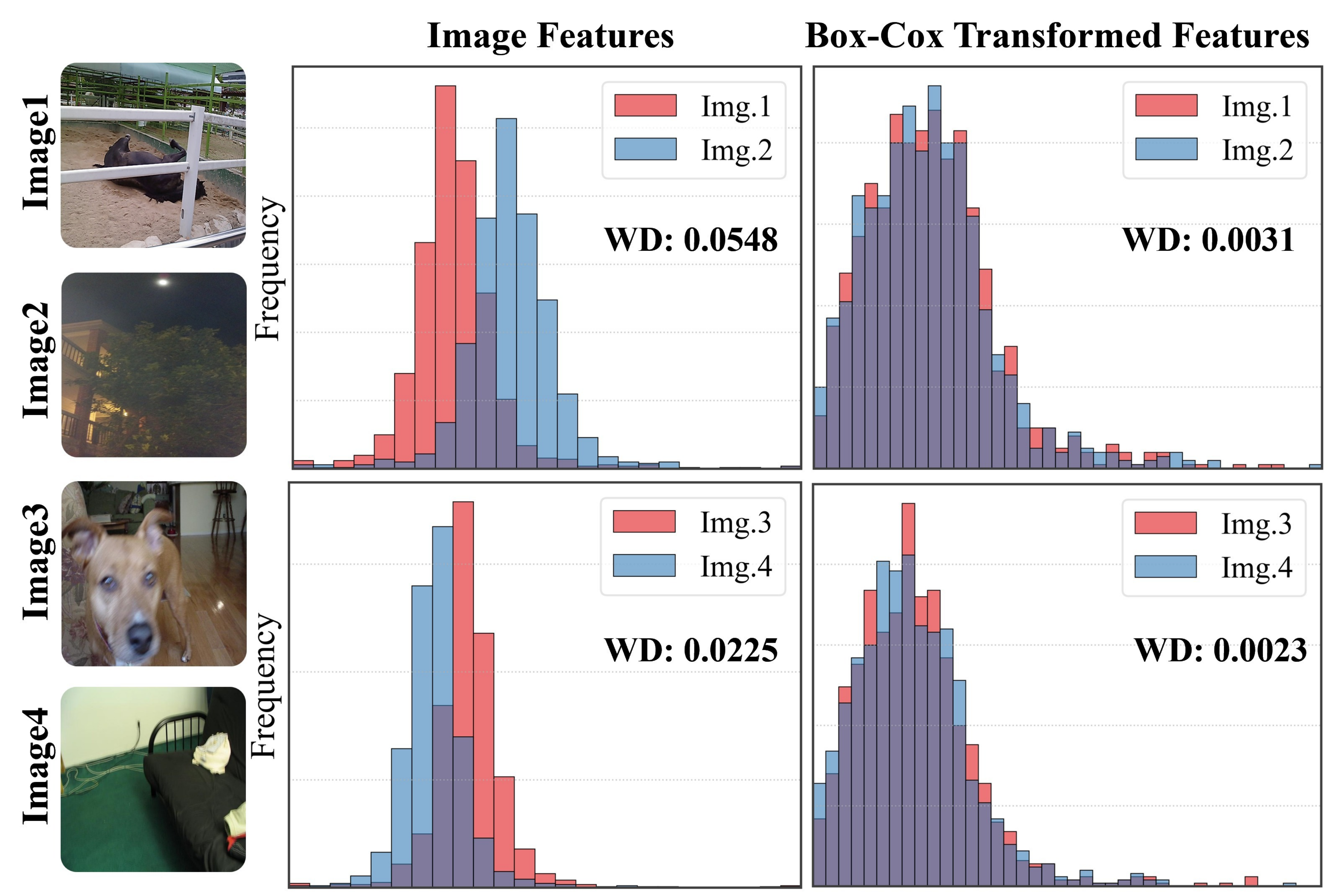}
    \caption{Semantic Bias exists in CLIP feature. This feature magnitudes for visually similar-quality images differ substantially across semantic categories. Statistical normalization is vital to make magnitude cues reliable. WD represents the Wasserstein Distance between two Feature distribution.}
    \label{fig:semantic_bias}
\end{figure}

As depicted in Eqn.~(\ref{eqn:cos}), the cosine similarity inherently normalizes both input vectors, which discards the magnitude information of the image embedding. However, we empirically observe that the magnitude also presents a high correlation with perceptual quality: High-quality images typically yield rich and discriminative features, reflected by larger magnitudes, while heavily degraded images exhibit reduced embedding norms. Nevertheless, as shown in Fig.~\ref{fig:semantic_bias}, the magnitude distributions vary significantly across different image content, even under similar perceptual quality. This content-dependent variation introduces a semantic bias that impairs direct comparison of magnitude scores across samples.

\subsubsection{Statistical Normalization via Box-Cox Transformation}
To mitigate the semantic bias inherent in raw CLIP embeddings, we introduce a statistical normalization approach based on the Box-Cox transformation~\cite{box1964analysis}, which is a classical power transform that stabilizes variance and reduces distributional skewness. Unlike cosine similarity, which discards magnitude information via $\ell_2$ normalization, our goal is to retain and standardize this information for quality prediction. Given the image-level feature embedding ${F}_{{img}} \in \mathbb{R}^D$, we first take its element-wise absolute value to remove the polarity and retain only activation strength:
 \begin{equation}
\hat{{F}} = |{F}_{{img}}| \in \mathbb{R}^D.
\end{equation}
Then we  normalize it by its standard deviation:
\begin{equation}
\tilde{{F}} = \frac{\hat{{F}} }{\sigma + \varepsilon},
\end{equation}
where $\sigma$ is the standard deviation computed over all dimensions, and $\varepsilon$ is a small constant for numerical stability. The Box-Cox transformation is finally applied independently to each dimension:
\begin{equation}
\mathbf{T}_d =
\begin{cases}
\dfrac{(\tilde{{F}}_d + 1)^\lambda - 1}{\lambda}, & \lambda \ne 0, \\
\log(\tilde{{F}}_d + 1), & \lambda = 0,
\end{cases}
\end{equation}
where $\mathbf{T}_d$ denotes the transformed $d$-th feature, and $\lambda$ is the power parameter. Finally, the normalized magnitude-based quality score is obtained by averaging across all dimensions:
\begin{equation}
Q_{{mag}} = \frac{1}{D} \sum_{d=1}^{D} \mathbf{T}_d.
\end{equation}
This transformation yields a statistically normalized scalar that effectively captures perceptual quality variations while remaining robust to semantic content.

\subsubsection{Confidence-Guided Adaptive Fusion}
\label{subsubsec:fusion}
While both $Q_{{sim}}$ and $Q_{{mag}}$ offer valuable but distinct quality cues, their reliability is not uniform across different image conditions. Specifically, $Q_{{sim}}$, which relies on semantic similarity between image embedding and textual prompts, is more robust when semantic content is well-preserved, such as in high-quality images. In contrast, $Q_{{mag}}$ captures distortions through statistical deviations in embedding magnitude, making it more responsive under severe degradation where semantic alignment is compromised.

\noindent To adaptively leverage their complementary strengths, we design a fusion scheme that dynamically adjusts the contribution of each cue based on their agreement. We begin by computing the discrepancy between the two estimates:
\begin{equation}
    \Delta = Q_{{sim}} - Q_{{mag}},
\end{equation}
which quantifies the direction and degree of disagreement. A large positive $\Delta$ suggests that $Q_{{sim}}$ is relatively confident (\eg, in a clean image), whereas a negative $\Delta$ indicates greater reliability in $Q_{{mag}}$ (\eg, when content is distorted). This discrepancy implicitly reflects the underlying quality level of the image and serves as a signal for confidence reweighting. We convert $\Delta$ into two fusion logits through an affine transformation:
\begin{align}
\label{eqn:fusion}
    \gamma_{{sim}} &= 1.0 + \alpha \Delta, \\
    \gamma_{{mag}} &= 0.6 - \alpha \Delta,
\end{align}
where $\alpha$ is a tunable hyperparameter controlling the sensitivity of the fusion to confidence gaps. The base constants (1.0 and 0.6) encode prior trust in the two metrics, while $\Delta$ adaptively adjusts these values based on content quality. We then apply softmax normalization to ensure the resulting weights form a valid probability distribution:
\begin{equation}
    [w_{{sim}}, w_{{mag}}] = \text{softmax}([\gamma_{{sim}}, \gamma_{{mag}}]).
\end{equation}
Finally, the overall perceptual quality score is obtained as a convex combination of both cues:
\begin{equation}
    Q = w_{{sim}} \cdot Q_{{sim}} + w_{{mag}} \cdot Q_{{mag}}.
\end{equation}

\section{Experiments}

\begin{table*}[t]
\centering
\small  
\begin{tabular}{@{}cl@{\hspace{1.5mm}}|@{\hspace{1.5mm}}*{9}{c}@{\hspace{1.5mm}}|@{\hspace{1.5mm}}l@{}}
\toprule
\multirow{2}{*}{\textbf{}} & \multirow{2}{*}{\textbf{Dataset}} 
& \multicolumn{10}{c}{\textbf{Method}} \\
\cmidrule(lr){3-12}
& & \textbf{NIQE} & \textbf{QAC} & \textbf{PIQE} & \textbf{dipIQ} & \textbf{SNP-NIQE} & \textbf{NPQI} & \textbf{ContentSep} & \textbf{MDFS} & \textbf{CLIP-IQA} & \textbf{MA-CLIP} \\
\midrule

\multirow{7}{*}{\rotatebox[origin=c]{90}{\textbf{SRCC}}} 
& CLIVE & 0.4495 & 0.2258 & 0.2325 & 0.2089 & 0.4654 & 0.4752 & 0.5060 & 0.4821 & 0.7019 &  \textbf{0.7428} \scriptsize {\textbf{({+5.8\%})}} \\
& CSIQ & 0.6191 & 0.4804 & 0.5120 & 0.5191 & 0.6090 & 0.6341 & 0.5871 & \textbf{0.7774} & 0.6807 &  0.7374 \scriptsize \textbf{({+8.3\%})} \\
& TID2013 & 0.3106 & 0.3719 & 0.3636 & 0.4377 & 0.3329 & 0.2804 & 0.2530 & 0.5363 & 0.5786 &  \textbf{0.5990} \scriptsize \textbf{({+3.5\%})} \\
& KADID & 0.3779 & 0.2394 & 0.2372 & 0.2977 & 0.3719 & 0.3909 & 0.5060 & \textbf{0.5983} & 0.5009 &  0.5251 \scriptsize \textbf{({+4.8\%})} \\
& KonIQ & 0.5300 & 0.3397 & 0.2452 & 0.2375 & 0.6284 & 0.6132 & 0.6401 & 0.7333 & 0.6846 &  \textbf{0.7645} \scriptsize \textbf{({+11.7\%})} \\
& SPAQ & 0.3105 & 0.4397 & 0.2317 & 0.2189 & 0.5402 & 0.5999 & 0.7084 & 0.7408 & 0.7144 &  \textbf{0.7725} \scriptsize \textbf{({+8.1\%})} \\
\cmidrule(lr){2-12}
& AVG & 0.4329 & 0.3495 & 0.3037 & 0.3200 & 0.4913 & 0.4990 & 0.5682 & 0.6487 & 0.6296 &  \textbf{0.6902} \scriptsize \textbf{({+9.6\%)}} \\

\midrule    

\multirow{7}{*}{\rotatebox[origin=c]{90}{\textbf{PLCC}}} 
& CLIVE & 0.4939 & 0.2841 & 0.3144 & 0.3163 & 0.5199 & 0.4920 & 0.5130 & 0.5364 & 0.7217 &  \textbf{0.7680} \scriptsize \textbf{(+6.4\%)} \\
& CSIQ & 0.6901 & 0.5934 & 0.6279 & 0.7009 & 0.6962 & 0.6479 & 0.3632 & \textbf{0.7907} & 0.7270 &  0.7828 \scriptsize \textbf{(+7.7\%)} \\
& TID2013 & 0.3789 & 0.4190 & 0.4615 & 0.4746 & 0.4055 & 0.4000 & 0.2203 & 0.6242 & 0.6552 &  \textbf{0.6756} \scriptsize \textbf{(+3.1\%)} \\
& KADID & 0.3883 & 0.3088 & 0.2887 & 0.3832 & 0.4212 & 0.3401 & 0.3568 & \textbf{0.5939} & 0.5204 &  0.5489 \scriptsize \textbf{(+5.5\%)} \\
& KonIQ & 0.4835 & 0.2906 & 0.2061 & 0.3773 & 0.6222 & 0.6139 & 0.6274 & 0.7123 & 0.7124 &  \textbf{0.8035} \scriptsize \textbf{(+12.8\%)} \\
& SPAQ & 0.2639 & 0.4497 & 0.2488 & 0.2239 & 0.5469 & 0.6155 & 0.6648 & 0.7177 & 0.7179 &  \textbf{0.7775} \scriptsize \textbf{(+8.3\%)} \\
\cmidrule(lr){2-12}
& AVG & 0.4498 & 0.3909 & 0.3579 & 0.4127 & 0.5353 & 0.5182 & 0.4576 & 0.6625 & 0.6981 &  \textbf{0.7261} \scriptsize \textbf{(+4.0\%)} \\
\bottomrule
\end{tabular}
\caption{Performance comparison of opinion-unaware IQA models on six benchmark datasets. The best results are \textbf{bolded}. Relative gains of MA-CLIP over CLIP-IQA are annotated in each cell.}
\label{tab:overall_optimized}
\end{table*}

\subsection{Experimental Settings}

\subsubsection{Implementation Details}
We implement our method using PyTorch, based on the pretrained Resnet50 model. All evaluations are conducted under a zero-shot setting, where no ground-truth supervision is used during model optimization. For the Box-Cox transformation in $Q_{{mag}}$, the power parameter $\lambda$ is empirically set to $0.5$, and we add $1.0$ as an offset to ensure positivity.  The  $\alpha$ in Eqn.~(\ref{eqn:fusion}) is fixed by $1.0$. All experiments are conducted using a single NVIDIA 3090 GPU.

\subsubsection{Datasets}
We evaluate our method on a wide range of datasets to verify its generalization and robustness. These include:
\textbf{(1) Synthetic distortion datasets}: CSIQ~\cite{larson2010most}, TID2013~\cite{ponomarenko2015image}, and KADID-10k~\cite{lin2019kadid}, containing various artificially generated distortions.
\textbf{(2) Authentic distortion datasets}: CLIVE~\cite{ghadiyaram2015massive}, KonIQ~\cite{hosu2020koniq}, and SPAQ~\cite{fang2020perceptual}, reflecting real-world degradations from mobile photography.
\textbf{(3) Image restoration (IR) datasets}: PIPAL~\cite{jinjin2020pipal}, featuring restored images from SR/denoising/deblurring pipelines.
\textbf{(4) AIGC quality datasets}: AGIQA-1k~\cite{zhang2023perceptual} and AGIQA-3k~\cite{li2023agiqa}, designed for quality assessment of AI-generated content.
All evaluations follow the standard protocol, where Spearman's Rank Correlation Coefficient (SRCC) and Pearson Linear Correlation Coefficient (PLCC) are reported.

\subsubsection{Comparison Methods}

We compare our method with two categories of existing IQA approaches:
\textbf{(1) Opinion-Unaware (OU) Methods.} These methods do not require subjective opinion scores for training and are typically used in a zero-shot manner. We include NIQE~\cite{mittal2012making}, QAC~\cite{xue2013learning}, PIQE~\cite{venkatanath2015blind}, LPSI~\cite{wu2015highly}, dipIQ~\cite{ma2017dipiq}, SNP-NIQE~\cite{liu2019unsupervised}, NPQI~\cite{liu2020blind}, CLIPIQA~\cite{wang2023exploring}, ContentSep~\cite{babu2023no}, and MDFS~\cite{ni2024opinion} in this category. Since our method is also a zero-shot approach that requires no training on opinion scores, this group serves as the most appropriate baseline for a fair comparison. 
\textbf{(2) Learning-based Methods.}  These methods rely on training with human-annotated quality scores and are typically optimized for specific IQA datasets. We include Re-IQA~\cite{saha2023re}, ARNIQA~\cite{agnolucci2024arniqa}, CLIP-IQA$^{+}$~\cite{wang2023exploring}, and GRepQ~\cite{srinath2024learning}. 
For a comprehensive evaluation of generalization capability, we compare with these learning-based methods trained on the training split of each testing dataset.

\begin{table}[t]
  \centering
  \footnotesize
  \renewcommand{\arraystretch}{1.1}
  \setlength{\tabcolsep}{0.6mm}
  \begin{tabular}{l|cc|cccccc}
  \toprule
  \multirow{2}{*}{{Method}} & \multicolumn{2}{c|}{{Setting}} & \multicolumn{2}{c}{PIPAL} & \multicolumn{2}{c}{AGIQA-1k} & \multicolumn{2}{c}{AGIQA-3k} \\
  \cmidrule(lr){2-3} \cmidrule(lr){4-5} \cmidrule(lr){6-7} \cmidrule(lr){8-9}
   & CD & ZS & \srcc & \plcc & \srcc & \plcc & \srcc & \plcc \\
  \midrule
  Re-IQA  & $\checkmark$ & & 0.568 & 0.587 & 0.783 & 0.840 & 0.811 & 0.874 \\ 
  ARNIQA & $\checkmark$ & & 0.634 & 0.666 & 0.768 & 0.849 & 0.803 & 0.881 \\ 
  CLIP-IQA$^{+}$ & $\checkmark$ & & 0.552 & 0.558 & 0.817 & 0.855 & 0.844 & 0.894 \\ 
  GRepQ & $\checkmark$ & & 0.554 & 0.568 & 0.740 & 0.797 & 0.807 & 0.858 \\ 
  \midrule
  NIQE & & $\checkmark$ & 0.167 & 0.181 & 0.623 & 0.721 & 0.510 & 0.526 \\ 
  IL-NIQE & & $\checkmark$ & 0.231 & 0.220 & \textbf{0.645} & \textbf{0.757} & 0.528 & 0.544 \\ 
  CL-MI & & $\checkmark$ & 0.281 & 0.282 & 0.474 & 0.621 & 0.591 & 0.665 \\ 
  CLIP-IQA & & $\checkmark$ & 0.332 & 0.339 & 0.511 & 0.644 & 0.658 & 0.716 \\ 
  \midrule
  \textbf{MA-CLIP} & & $\checkmark$ & \textbf{0.371} & \textbf{0.393} & 0.528 & \textbf{0.668} & \textbf{0.706} & \textbf{0.764} \\ 
  \textbf{Gain} & & & \rotatebox{0}{\scriptsize \textbf{+11.7\%}} 
  & \rotatebox{0}{\scriptsize \textbf{+15.9\%}} 
  & \rotatebox{0}{\scriptsize \textbf{+3.3\%}} 
  & \rotatebox{0}{\scriptsize \textbf{+3.9\%}} 
  & \rotatebox{0}{\scriptsize \textbf{+7.3\%}} 
  & \rotatebox{0}{\scriptsize \textbf{+6.7\%}} \\
  \bottomrule
  \end{tabular}
  \caption{Quantitative results on image restoration and AIGC datasets. CD: Cross-dataset, ZS: Zero-shot. The last row shows the relative gain of MA-CLIP compared to CLIP-IQA.  Best scores are highlighted in bold.}
  \label{tab:restoration_aigc}
\end{table}

\subsection{Quantitative Comparison}

As summarized in Table~\ref{tab:overall_optimized}, our Magnitude-Aware CLIP (MA-CLIP) achieves consistent and significant performance gains compared with CLIP-IQA across all benchmark categories, demonstrating its robustness under various distortion types and domains.

On synthetic distortion datasets of CSIQ, our method outperforms existing opinion-unaware baselines and CLIP-based models by a large margin. For instance, on TID2013, MA-CLIP achieves an SRCC of 0.599, surpassing CLIP-IQA by 3.5\%. This improvement reflects the benefit of incorporating distortion-sensitive features via our normalized magnitude modeling, which enhances the model’s ability to detect subtle degradation patterns often present in synthetic benchmarks. 
On real-world datasets like KonIQ-10k and SPAQ, where distortions are more diverse and semantically entangled, MA-CLIP achieves an SRCC of 0.765 on KonIQ-10k, outperforming all OU methods. This shows that the combination of $Q_{{sim}}$ and statistically normalized $Q_{{mag}}$ enables a more holistic perception of quality.

To further test generalization in downstream applications, Table~\ref{tab:restoration_aigc} reports results on image restoration (PIPAL) and AIGC (AGIQA-1k/3k) datasets. These datasets feature challenging distribution shifts, such as hallucination artifacts, over-smoothing, or texture inconsistency, which are often poorly handled by traditional or purely semantic-based metrics. MA-CLIP achieves the best PLCC of 0.706 and SRCC of 0.764 on AGIQA-3k, surpassing recent multimodal methods like MDFS. It also performs competitively on PIPAL, where many models struggle due to the diverse restoration algorithms and overfitting risks.

The performance gain confirms the advantage of our adaptive fusion design, which dynamically balances semantic and magnitude cues based on image-specific confidence. In addition, we compare against learning-based methods trained on KonIQ-10k (\eg, GRepQ, CLIP-IQA$^+$). While these models leverage large-scale opinion-aware data, our zero-shot MA-CLIP still achieves highly competitive results, especially on datasets it has never seen during training. This highlights the strong generalization capability of our method, without sacrificing interpretability or requiring expensive annotations.

\begin{figure}[t]
    \centering
    \includegraphics[width=\linewidth]{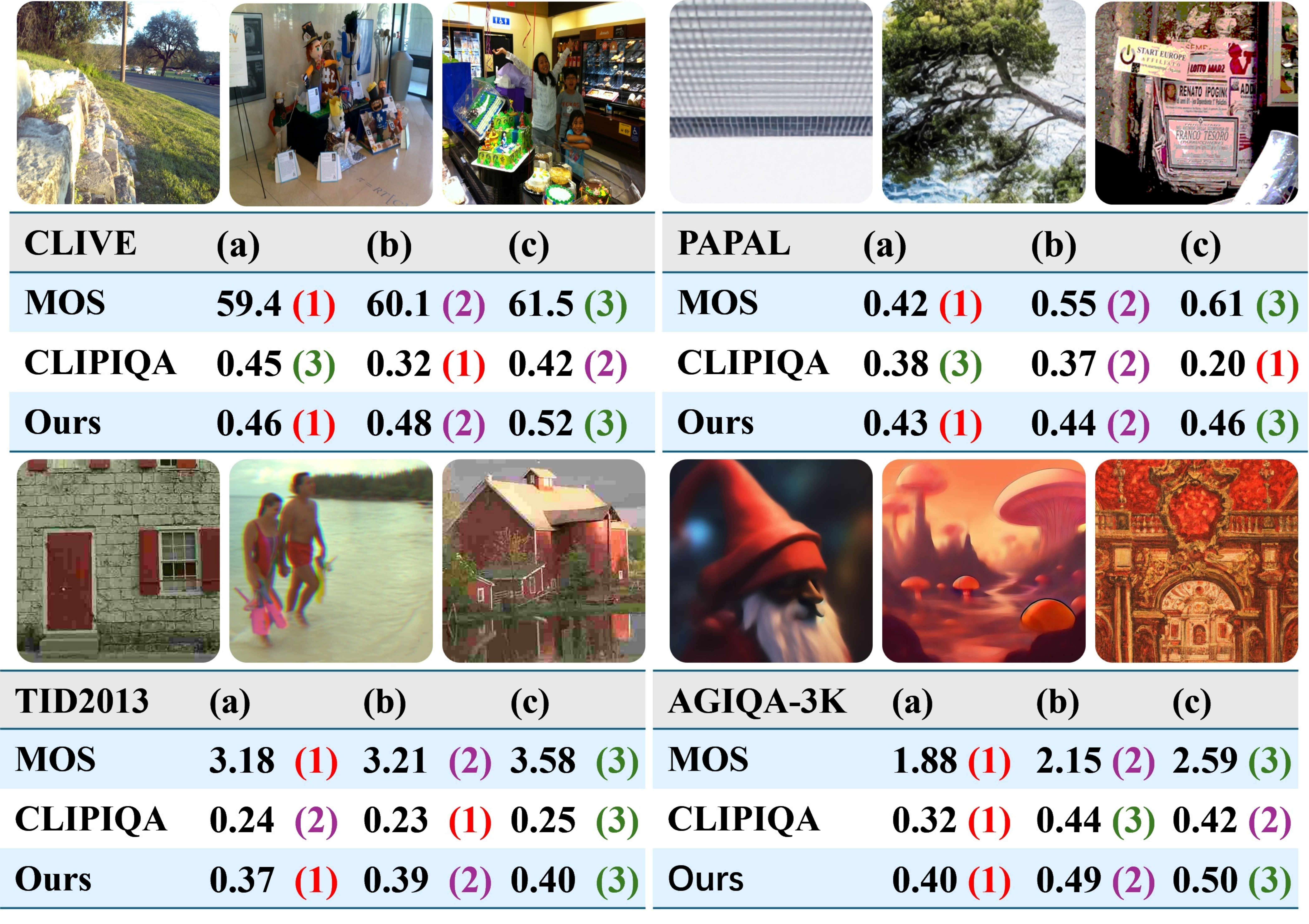}
   \caption{MOS alignment visualization. Representative examples from multiple datasets illustrating the ranking alignment between MOSs and our MA-CLIP predictions. Each triplet shows three images with their MOSs and predicted quality scores of CLIP-IQA and our  MA-CLIP. }
    \label{fig:mos_vis}
\end{figure}
\subsection{Qualitative Visualization}

To further complement the quantitative results, we provide two qualitative visualization that illustrate the efficacy of our proposed method from a human-understandable perspective:

\paragraph{(1) MOS alignment visualization} As shown in Fig.~\ref{fig:mos_vis}, we present representative examples from diverse datasets where the original CLIP-IQA model fails to correctly rank the image qualities in accordance with the MOS. In each set, we display three images along with their  MOS and the predicted quality scores from both CLIP-IQA and our proposed MA-CLIP. It can be observed that CLIP-IQA often over-relies on semantic content and yields inverted or inconsistent rankings. In contrast, our method incorporates magnitude-aware correction, which adjusts the quality estimates to better reflect perceptual degradation. As a result, the predicted order aligns more closely with the MOS-based ranking.

\begin{figure}[t]
    \centering
    \includegraphics[width=\linewidth]{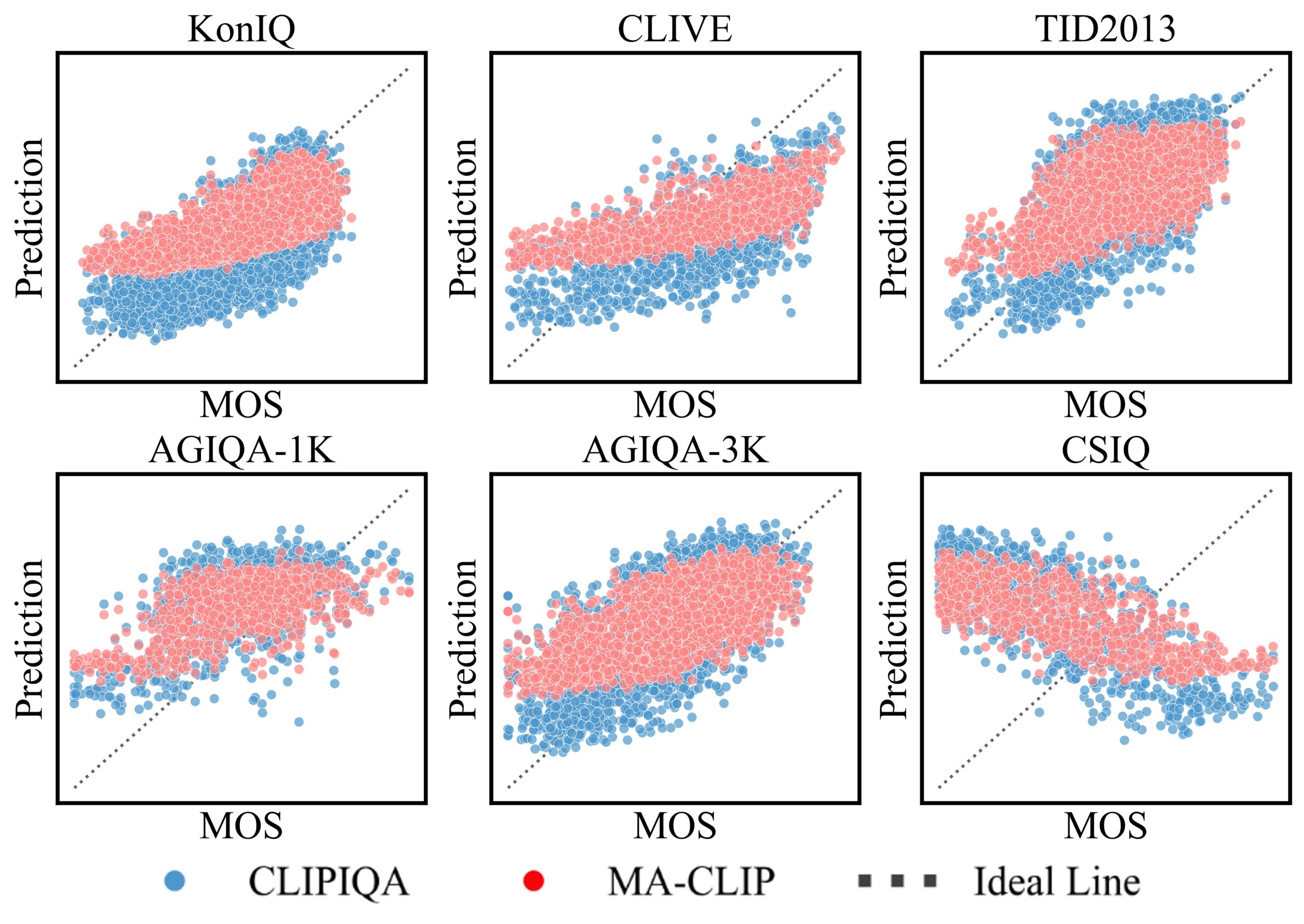}
   \caption{Scatter plot comparison of CLIP-IQA and MA-CLIP. The x-axis represents the MOS, while the y-axis shows the prediction. As the scatter gets closer to the ideal line, it indicates that the model predicts better.}
    \label{fig:scatter_comparison}
\end{figure}

\paragraph{(2) Scatter plot comparison} In Fig.~\ref{fig:scatter_comparison}, we visualize the scatter plots comparing CLIP-IQA and MA-CLIP over six datasets including four representative groups: (a) synthetic distortions, (b) authentic distortions, (c) IR (mage restoration, and (d) AIGC-generated images. 
Each plot maps the predicted scores versus the  MOSs, where the ideal prediction would lie along the diagonal line. 
The scatter patterns reveal that CLIP-IQA suffers from more dispersed and biased predictions, especially on complex or underrepresented distortions. In contrast, MA-CLIP exhibits tighter clustering around the diagonal, indicating improved consistency and robustness in ranking. 

\subsection{Ablation Study}

To thoroughly validate the effectiveness and design choices of our proposed MA-CLIP framework, we conduct a series of ablation experiments. These experiments are designed to isolate and quantify the contribution of each key component, including individual prediction branches, magnitude feature extraction strategies, fusion mechanisms, statistical parameters, and backbone architectures. 

\paragraph{(1) Contribution of Each Branch.}
We first evaluate the independent performance of the two scoring branches: $Q_{{sim}}$, which captures semantic similarity between image and text embeddings by CLIP model; and $Q_{{mag}}$, which estimates distortion severity via magnitude information. As reported in Table~\ref{tab:ablation_branch}, $Q_{{sim}}$ performs well on content-consistent distortions but fails to capture quality degradation in texture-corrupt or over-smoothed cases, often overemphasizing high-level semantics. In contrast, $Q_{{mag}}$ exhibits stronger sensitivity to signal-level degradations but is less reliable when semantic preservation is critical. The proposed confidence-guided fusion of both branches results in a substantial SRCC gain over 8.5\% , demonstrating their complementary nature.

\begin{table}[t]
\centering
\begin{tabular}{@{}l*{7}{@{\hspace{2mm}}c}@{}} 
\toprule
\multirow{2}{*}{\textbf{Dataset}} & \multicolumn{2}{c}{\textbf{$Q_{{sim}}$}} & \multicolumn{2}{c}{\textbf{$Q_{{mag}}$}} & \multicolumn{2}{c}{\textbf{Fusion}} \\
\cmidrule(lr){2-3}\cmidrule(lr){4-5}\cmidrule(lr){6-7}
 & SRCC & PLCC & SRCC & PLCC & SRCC & PLCC & \\
\midrule
CLIVE   & 0.702 & 0.722 & 0.418 & 0.503 & \textbf{0.743} & \textbf{0.768}\\
CSIQ    & 0.681 & 0.727 & 0.448 & 0.460 & \textbf{0.737} & \textbf{0.783}\\
KonIQ-10k   & 0.685 & 0.712 & 0.557 & 0.616 & \textbf{0.765} & \textbf{0.804} \\
SPAQ    & 0.714 & 0.718 & 0.578 & 0.598 & \textbf{0.773} & \textbf{0.775} \\
\midrule
AVG & 0.696 & 0.720 & 0.500 & 0.544 & \textbf{0.755} & \textbf{0.783}\\
\bottomrule
\end{tabular}
\caption{Ablation study on the contribution of each scoring branch. 
         $Q_{{sim}}$: semantic-only; $Q_{{mag}}$: magnitude-only; 
         Fusion: confidence-guided combination.}
\label{tab:ablation_branch}
\end{table}

\paragraph{(2) Variants of Magnitude Feature Extraction.}
We compare three strategies for computing the magnitude-based feature: (i) L1 norm, (ii) L2 norm, and (iii) our proposed Box-Cox normalized features norm. As shown in Table~\ref{tab:ablation_norm}, the Box-Cox-based normalization yields consistently higher correlation with perceptual quality scores across all datasets. In particular, it improves SRCC by 59\% over the L2 variant on KonIQ-10k, highlighting the benefits of distributional stabilization. 
When each variant is used independently (without fusion), Box-Cox normalization again outperforms the alternatives, indicating its standalone robustness in capturing distortion-aware cues.

\begin{table}[t]
    \centering
    \begin{tabular}{@{}l *{6}{c}@{}}
        \toprule
        \multirow{2}{*}{\textbf{Type}} 
        & \multicolumn{2}{c}{\textbf{CSIQ}} & \multicolumn{2}{c}{\textbf{CLIVE}} 
        & \multicolumn{2}{c@{}}{\textbf{KonIQ-10k}}  \\ 
        \cmidrule(lr){2-3}\cmidrule(lr){4-5}\cmidrule(lr){6-7}
        & SRCC & PLCC  & SRCC & PLCC & SRCC & PLCC  \\
        \midrule
        L1 & 0.126 & 0.152 & 0.345 & 0.369  & 0.495 & 0.530 \\
        L2 & 0.299 & 0.386 & 0.193 & 0.304  & 0.350 & 0.349 \\
        \midrule
        Ours & \textbf{0.448} & \textbf{0.461} & \textbf{0.418} & \textbf{0.503}  & \textbf{0.557} & \textbf{0.616} \\
        \bottomrule
    \end{tabular}
    \caption{Comparison on perceptual-optimized feature norm with different feature norm on CSIQ, CLIVE and KonIQ.}
    \label{tab:ablation_norm}
\end{table}

\paragraph{(3) Fusion Strategy Comparison.}
To assess the effectiveness of our confidence-weighted adaptive fusion, we compare it against several baselines: (i) equal-weighted average, and (ii) fixed-weight summations with various ratios. Results in Table~\ref{tab:ablation_fusion} show that adaptive fusion achieves superior performance across all tested datasets.   
This confirms that confidence-aware fusion dynamically adjusts to different image conditions, enabling better utilization of the complementary strengths of $Q_{{sim}}$ and $Q_{{mag}}$.
\begin{table}[t]
    \centering
    \begin{tabular}{@{}lcccccc@{}}
        \toprule
        \multirow{2}{*}{\textbf{Type}} 
        & \multicolumn{2}{c}{\textbf{CG-Fusion}}
        & \multicolumn{2}{c}{\textbf{CSIQ}} 
        & \multicolumn{2}{c@{}}{\textbf{KADID-10k}}  \\
        \cmidrule(lr){2-3}\cmidrule(lr){4-5}\cmidrule(lr){6-7}
        & $w_{{sim}}$ & $w_{{mag}}$  & SRCC & PLCC  & SRCC & PLCC  \\
        \midrule
        L1 & \multicolumn{2}{c}{$\checkmark$} & 0.126 & 0.152  & 0.394 & 0.397 \\
        L2 & \multicolumn{2}{c}{$\checkmark$} & 0.280 & 0.398  & 0.338 & 0.345 \\
        \midrule
        \multirow{4}{*}{Ours}  & 0.8 & 0.2 & 0.736 & 0.779  & 0.523 & 0.544 \\
         & 0.2 & 0.8 & 0.700 & 0.742  & 0.445 & 0.492 \\
         & 0.5 & 0.5 & 0.734 & 0.779  & 0.520 & 0.545 \\
         & \multicolumn{2}{c}{$\checkmark$} & \textbf{0.737} & \textbf{0.783}  & \textbf{0.525} & \textbf{0.549} \\
        \bottomrule
    \end{tabular}
    \caption{Ablation study on weight combination with different feature norm on CLIVE and KADID-10k.}
    \label{tab:ablation_fusion}
\end{table}

\begin{figure}[t]
    \centering
    \includegraphics[width=1.0\linewidth]{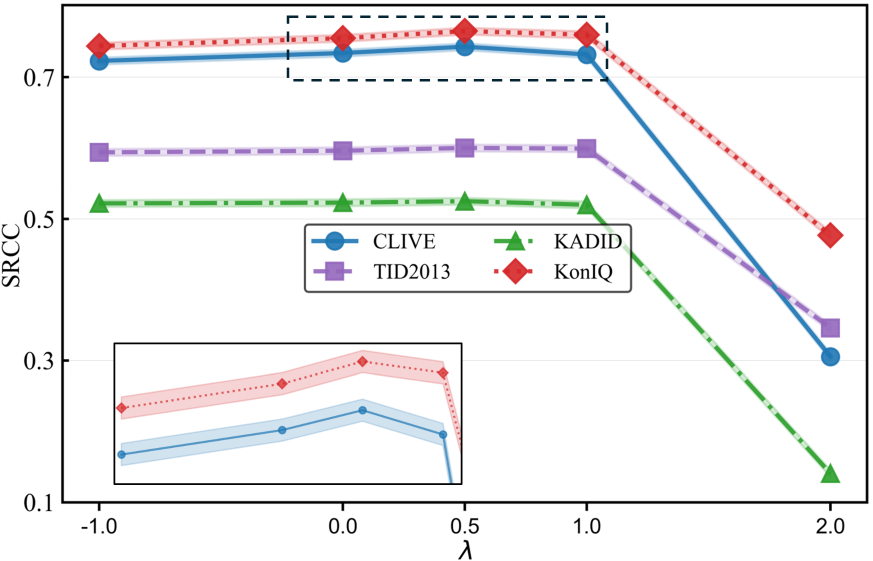}
   \caption{Sensitivity of Box-Cox parameter $\lambda$ on SRCC across on CLIVE, TID2013, KonIQ-10k and KADID-10k.}
    \label{fig_lambda}
\end{figure}

\paragraph{(4) Sensitivity to Box-Cox Transformation Parameter $\lambda$.}
To investigate the robustness of the Box-Cox normalization, we conduct a sensitivity analysis on the transformation parameter $\lambda$, which controls the degree of non-linearity applied to magnitude values. As plotted in Fig.~\ref{fig_lambda}, we observe that small positive values (\eg, $\lambda=0.5$) consistently yield stable and high SRCC values. Very large  $\lambda$ leads to a performance drop due to either over-flattening (loss of signal variance) or numerical instability. These findings suggest that light nonlinear normalization is sufficient to suppress magnitude outliers while preserving distortion-relevant information.

\paragraph{(5) Impact of Backbone Architectures.}
We further assess the generality of MA-CLIP across diverse CLIP backbones, including ResNet-50, ResNet-101, ViT-B/32, and ViT-L/14. As shown in Table~\ref{tab:backbone}, our magnitude-aware design consistently improves performance across all architectures, surpassing the corresponding CLIP-IQA baselines in both SRCC and PLCC. The consistent gains highlight the effectiveness of magnitude-aware correction even when applied to models with high semantic representation capacity.

\begin{table}[t]
  \centering
  \begin{tabular}{@{}cl*{6}{@{\hspace{1.2mm}}c}@{}}
  \toprule
  & \multirow{2}{*}{ \textbf{Method}} & \multicolumn{2}{c}{\textbf{CSIQ}} & \multicolumn{2}{c}{\textbf{KonIQ-10k}} & \multicolumn{2}{c}{\textbf{AGIQA-3k}}\\
  \cmidrule(lr){3-4}\cmidrule(lr){5-6}\cmidrule(lr){7-8}
  & & SRCC & PLCC & SRCC & PLCC & SRCC & PLCC  \\
  \midrule
  \multirow{2}{*}{\rotatebox[origin=c]{90}{R50}}
  & CLIPIQA & 0.681 & 0.727 & 0.685 & 0.712 & 0.658 & 0.716  \\
  &\textbf{Ours}
  & \textbf{0.737} & \textbf{0.783}
  & \textbf{0.765} & \textbf{0.804}
  & \textbf{0.706} & \textbf{0.764} \\
  \midrule
    \multirow{2}{*}{\rotatebox[origin=c]{90}{R101}}
  & CLIPIQA & 0.715 & 0.705 & 0.710 & 0.730 & 0.643 & 0.697  \\
  & \textbf{Ours}
  & \textbf{0.741} & \textbf{0.764}
  & \textbf{0.727} & \textbf{0.748}
  & \textbf{0.667} & \textbf{0.743} \\
  \midrule
    \multirow{2}{*}{\rotatebox[origin=c]{90}{B/32}}
  & CLIPIQA & 0.763 & 0.783 & 0.715 & 0.743 & 0.663 & 0.710  \\
  & \textbf{Ours}
  & \textbf{0.783} & \textbf{0.810}
  & \textbf{0.760} & \textbf{0.803}
  & \textbf{0.694} & \textbf{0.757} \\
  \midrule
  \multirow{2}{*}{\rotatebox[origin=c]{90}{L/14}}
  & CLIPIQA & 0.622 & 0.628 & 0.682 & 0.709 & 0.699 & 0.788  \\
  & \textbf{Ours}
  & \textbf{0.666} & \textbf{0.680}
  & \textbf{0.717} & \textbf{0.752}
  & \textbf{0.717} & \textbf{0.814} \\
  \bottomrule
  \end{tabular}
  \caption{Impact of different CLIP backbones on MA-CLIP performance (SRCC/PLCC). Best scores for each backbone are highlighted in bold.}
  \label{tab:backbone}
\end{table}

\section{Conclusion}
In this work, we revisit CLIP-based NR-IQA by identifying a crucial yet previously overlooked quality cue: the magnitude of CLIP image features. While existing CLIP-IQA approaches rely solely on prompt-based cosine similarity, we demonstrate that feature magnitude exhibits strong and complementary correlation with perceptual quality. To harness both cues effectively, we propose a novel, training-free dual-source framework that integrates a statistically normalized magnitude score with semantic similarity via a confidence-guided fusion strategy. Extensive experiments across diverse IQA benchmarks show that our method consistently outperforms both CLIP-IQA and state-of-the-art NR-IQA models, without requiring any task-specific fine-tuning. These findings highlight the value of revisiting internal properties of pretrained models and open new directions for plug-and-play quality assessment leveraging multimodal embeddings.

\section{Acknowledgments}
The work was supported by the National Natural Science Foundation of China under Grant No. 62401214, the National Natural Science Foundation of China under Grant No. 62477015, the Key Research and Development Program of Guangdong of China under Grant No. 2023B0303010004, and the Innovation Team Project for Universities in Guangdong Province in China under Grant No. 2023KCXTD011.

\appendix

\bigskip

\bibliography{CameraReady/LaTeX/aaai2026_red}

\end{document}